\documentclass[10pt,journal]{IEEEtran}

\usepackage[T1]{fontenc}
\usepackage[utf8]{inputenc}
\usepackage{amsmath,amssymb}
\usepackage{graphicx}
\usepackage{booktabs}
\usepackage{array}
\usepackage[hidelinks]{hyperref}
\usepackage{url}
\usepackage{xurl}
\usepackage{microtype}

\newcommand{\FM}{\mathrm{FM}}
\newcommand{\dev}{\mathrm{dev}}
\newcommand{\R}{\mathcal{R}}

\newcommand{\repourl}{https://github.com/arjunbhupatiraju/cns-pns-regeneration}

\begin{document}

\title{Predictive Multiplicity in Cell-Fate Assignment:
Label-Free Rashomon Sets and the Limits of Per-Cell Certification}

\author{Arjun~Bhupatiraju\\[2pt]
\normalfont\itshape McNeil High School, Austin, Texas\\[1pt]
\normalfont ORCID: 0009-0006-2667-0098\\[1pt]
\normalfont\small arjun.bhupatiraju28@gmail.com}

\maketitle

\begin{abstract}
Single-cell trajectory inference maps transcriptomic measurements onto
developmental continua, yet parameter configurations that fit the data with
equivalent statistical fidelity can assign conflicting cell fates. This
paper presents FateMultiplicity, a label-free framework that constructs a
statistically admissible model set, or Rashomon set, without ground-truth
lineage labels, by evaluating model discrepancy on cross-fitted held-out
genes under non-inferiority testing calibrated against random-seed
variation. Multiplicity so measured is large, and depends more on the
diversity of the model space than on its size: twelve configurations of a
second algorithm expose 20.0\% of cells where twenty-four of the first
expose 3.8\%. Whether the resulting per-cell certified fate margin $\FM$
yields more reliable assignments than the fitted model already provides is
then tested, and it does not. On simulation ground truth, with every
statistic computed on the same cells, $\FM$ discriminates misassignment at
AUC 0.682 against 0.965 for the baseline configuration's own decision
margin ($p = 0.003$) and 0.854 for a seed-dispersion baseline needing no
acceptance region. The failure is mechanistic and specific to the lower
tail. Informativeness is governed by the breadth of the admitted set rather
than its cardinality: at fixed cardinality four, seed refits give 0.933 and
hyperparameter-perturbed sets 0.701. Relaxing the infimum to a $q$-quantile
recovers discrimination monotonically but converges toward the single
model's own confidence rather than exceeding it, and the supremum over the
same margins reaches 0.973 because $\theta^*$'s own membership bounds it
from below. The infimum alone is unanchored, free to report whichever
admitted configuration disagrees most. Multiplicity in trajectory inference
is therefore worth measuring and reporting, while per-cell certification
over a label-free Rashomon set is not a route to more reliable fate calls.
Two constructions survive: a margin-erosion ratio separates real from
spurious branch points in simulation (AUC 0.890, untested on real data)
where certification cannot, and against
clonally observed fate, uncertified cells disagree with their clone's
outcome 16.4 percentage points more often than certified cells
($p < 0.001$).
\end{abstract}

\begin{IEEEkeywords}
Cell fate, model multiplicity, negative result, Rashomon set,
reproducibility, single-cell RNA sequencing, trajectory inference
\end{IEEEkeywords}

\IEEEpeerreviewmaketitle

\section{Introduction}

\noindent Single-cell RNA sequencing (scRNA-seq) has transformed
developmental biology by enabling the reconstruction of continuous cellular
differentiation trajectories from snapshot transcriptomic data. Trajectory
inference methods order individual cells along pseudotemporal axes and
compute fate probabilities to identify lineage commitment points
\cite{lange,setty,haghverdi}. However, these inference pipelines rely on
complex chains of algorithmic steps---dimensionality reduction,
nearest-neighbor graph construction, pseudotime ordering, and absorbing
random walks \cite{kemeny}---each governed by parameters the analyst
selects.

\subsection{Related Work and Background}

Multiple distinct models achieving near-identical performance while making
conflicting predictions on individual points is the Rashomon effect, or
predictive multiplicity \cite{breiman,fisher,marx,semenova}. Marx
\emph{et al.} \cite{marx} and Watson-Daniels \emph{et al.} \cite{wd}
established formal definitions in classification, with metrics for decision
ambiguity across candidate model sets; later frameworks address certified
robustness and uncertainty decomposition \cite{hsu,hullermeier}, isolating
model-driven variability from aleatoric noise.

In parallel, multiverse analysis evaluates pipelines by sweeping
combination spaces of processing parameters \cite{steegen,patel}, though
conventional approaches enumerate every combination without filtering
models that fit the data poorly. Trajectory benchmarking has documented
that different algorithms produce divergent topologies \cite{saelens}, but
with no formal notion of admissibility, separating legitimate variation
between configurations from outright model failure has remained difficult.

\subsection{The Methodological Gap}

Predictive multiplicity has been studied extensively in supervised
classification, where an explicit loss defines the admissible set.
Single-cell trajectory inference has no such loss: true fate labels are
rarely available, so the Rashomon set cannot be defined in the usual way.

FateMultiplicity resolves this by evaluating candidate configurations
against held-out genes with a cross-fitted discrepancy metric, admitting
models through non-inferiority testing calibrated against random-seed
noise. The decision-margin formulation and the interpretation of its sign
are adapted from Marx \emph{et al.} \cite{marx} and Watson-Daniels
\emph{et al.} \cite{wd}. The resulting per-cell quantity is the certified
fate margin ($\FM$), the minimum decision margin over all statistically
admissible configurations.

\subsection{What This Paper Reports}

Constructing the admissible set is the part that works, and the multiplicity
it exposes is large enough to matter for practice. Certifying individual
cells against it is the part that does not. This work set out to establish $\FM$ as
a per-cell reliability statistic and found instead that it is dominated by
the fitted model's own decision margin, and that the reason is structural
rather than particular to this pipeline: an infimum over a model set is an
extreme order statistic, and as the set broadens it reports the most extreme
admitted model instead of the cell. It is reported in full, with the diagnostics
that localize it to the aggregator rather than to the underlying quantity,
because the construction is an appealing one others are likely to attempt.

\section{Methodology}

\begin{table}[!t]
\caption{Summary of Formal Mathematical Notation}
\label{tab:notation}
\centering\footnotesize
\begin{tabular}{@{}ll@{}}
\toprule
Symbol & Definition \\
\midrule
$\Theta$ & Total hyperparameter search space \\
$\theta$, $\theta^*$ & Candidate and baseline configuration \\
$p_{i,k}(\theta)$ & Probability of cell $i$ committing to fate $k$ \\
$k^*$ & Primary baseline terminal fate assignment \\
$G_1^{(k)}$, $G_2^{(k)}$ & Training and held-out gene sets, fold $k$ \\
$K$ & Total cross-fitting folds ($K = 3$) \\
$d_g(\theta)$ & Cross-fitted discrepancy for gene $g$ \\
$\Delta_g(\theta)$ & $d_g(\theta) - d_g(\theta^*)$ \\
$\delta$ & Seed-calibrated margin ($1.086\times10^{-3}$) \\
$\alpha$ & Acceptance level, also the Benjamini--Hochberg \\
 & \quad false discovery rate across configurations \\
$\R(\alpha)$ & Rashomon set of admissible configurations \\
$\FM_i(\alpha)$ & Certified fate margin for cell $i$ \\
$\bar{m}_i(\alpha)$ & Companion supremum fate margin \\
\bottomrule
\end{tabular}
\end{table}

\subsection{Data Preprocessing and Compartment Selection}

Four datasets are used. The first is mouse spinal cord injury, GEO
GSE162610 \cite{milich}: 66{,}178 cells over four timepoints and two
dissociation protocols, from which the authors' annotations identify
19{,}818 microglia. After removing 118 high-ribosomal cells the analysis
was restricted to 10x v2 chemistry (15{,}627 cells), since v3 covered only
1\,dpi and uninjured, with dissociation protocol balanced across timepoints
and retained as a nuisance covariate. A sample-stratified subsample of
6{,}000 cells and 15{,}787 genes was used, the size being a memory
constraint on the dense matrices the discrepancy requires. Preprocessing
used Scanpy \cite{wolf}.

The second is GSE72857 \cite{paul}, 2{,}730 myeloid progenitors spanning an
established erythroid/myeloid bifurcation, used without subsampling. Two
contract parameters adjust for its size: the detectability floor holds its
fraction fixed at 3.3\%, and $\theta^*$ uses 1{,}500 highly variable genes
and 20 components. Its smaller gene universe yields 32 co-expression blocks
rather than 48, a coarser permutation null.

The third is the lineage-barcoded haematopoietic timecourse of Weinreb
\emph{et al.} \cite{weinreb}, where a heritable lentiviral barcode is read
out alongside the transcriptome, so the fate of a cell's descendants is
observed rather than inferred. A sample-stratified subsample of 12{,}000
cells was taken, state read at the early timepoint and fate from clone
members at the later ones. It is the only dataset carrying fate labels
formed independently of expression, which Section~\ref{sec:clonal} uses.

The fourth is the fibroblast-to-endoderm reprogramming timecourse of Biddy
\emph{et al.} \cite{biddy}, GSE99915: 11{,}999 cells and 19{,}268 genes
after joining counts to metadata on cell name. CellTag clone calls were not
obtainable in joinable form, so this arm reports multiplicity and
certification power, which need no labels, and not $\FM$ against observed
fate---its identity annotations derive from expression, so scoring a fate
margin against them would be circular.

\subsection{Gene Universe Partitioning and Feature Selection}

Feature selection isolates highly variable genes (HVGs) to construct the
trajectory, while held-out genes assess model discrepancy.

\begin{itemize}
\item \textbf{Baseline selection.} 6{,}000 HVGs were initially selected
 based on dispersion.
\item \textbf{Detectability floor.} Because dispersion-based selection
 inherently favors rare, highly skewed genes, a detectability floor
 requiring expression in $\ge 200$ cells was applied to establish the
 scoring universe. Unfiltered median gene detection was 56 cells of
 6{,}000; applying the floor increased median detection to 410 cells,
 retaining 1{,}660 of the 6{,}000 HVGs.
\item \textbf{Fold partitioning.} $K = 3$ fold cross-fitting was applied
 with a 20\% holdout fraction per fold, using disjoint fold assignments. A
 total of 990 genes were scored across all folds. This detectability floor
 adapts the standard minimum-cells-per-gene preprocessing filter,
 transferring its application from the model-fitting universe to the
 evaluation universe.
\end{itemize}

\subsection{Model Space and Grid Specification}

The baseline $\theta^*$ uses an absorbing random walk \cite{kemeny} solved
as a sparse linear system \cite{scipy}, with 2{,}000 HVGs, 30 principal
components, 15 nearest neighbors, seed 20260808, a backward step penalty of
0.05 and no teleportation. The walk's internal geodesic pseudotime is both
the ordering the discrepancy metric scores and the ordering driving the
transition operator, so admissibility selection acts directly on the
parameterization governing fate probabilities.

Grids span nine operational axes: HVG count, PC dimension, $k$-NN size,
seed, backward penalty, late-fraction threshold, terminal set size,
direction strength and self-loop weight, at levels fixed before execution
and enumerated in the repository's grid ledger. Perturbing single axes gives 24
configurations; perturbing pairs gives 100, of which 96 fit and 45 are
well-formed.

Thirteen Palantir configurations \cite{setty} supply a second algorithm,
varying gene count, components, graph construction, diffusion components,
waypoint density and seed. The gate needs no structural change for this,
since (1) scores ordering discrepancy regardless of what produced it.
Twelve yielded valid metrics; one configuration and one fold of another
were rejected by Palantir's internal check when branch probabilities failed
to sum to one, and both are retained in all denominators. Both adapters
report a \texttt{graph\_diffusion} provenance, so this spans two algorithms
within one family, and every multiplicity increment attributed to it below
is a lower bound on what a cross-family expansion would expose.

Because $\FM$ is an infimum over $\R(\alpha)$, adding configurations
expands the set and can only decrease or hold it constant, so reported
multiplicity rates are strict lower bounds tied to $|\Theta|$. And while
$\theta^*$ centres the acceptance region it is not optimal, nor even the
best-fitting member of its own admissible set: ten of the twelve Palantir
configurations achieve lower held-out discrepancy, the best reaching
$D = 0.98786$ against $0.9921$. Because (3) is one-sided, better-fitting
configurations are admitted without altering the construction.

\subsection{Cross-fitted Discrepancy and Non-inferiority Testing}

Model performance is evaluated without ground-truth labels by assessing the
expression smoothness of held-out genes along the inferred pseudotime
ordering. For a candidate configuration $\theta$ evaluated across $K$
folds,
\begin{equation}
d_g(\theta) = \frac{1}{K}\sum_{k=1}^{K} \dev_g\!\left(\theta;\, G_2^{(k)}\right),
\end{equation}
where $\theta$ is fitted strictly on training genes $G_1^{(k)}$ and
evaluated on held-out genes $G_2^{(k)}$. The single-gene deviance
$\dev_g(\theta; G_2)$ on held-out expression vector $x_g$ ordered by
pseudotime $i = 1 \dots n$ is
\begin{equation}
\dev_g(\theta; G_2) =
 \frac{\sum_{i=1}^{n-1}\left(x_{g,(i+1)} - x_{g,(i)}\right)^2}
 {2(n-1)\cdot \mathrm{Var}(x_g)}.
\end{equation}
Equation (2) is the von Neumann ratio \cite{vonneumann} and satisfies two
key formal properties: (i) its expected value under a random cell ordering
is exactly 1, and (ii) it depends strictly on the rank order of cells,
rendering it invariant to monotonic rescaling of pseudotime values.

In supervised learning a Rashomon set is a tolerance ball around the
empirical risk minimizer,
$\{\theta \in \Theta : L(\theta) \le L(\theta^*) + \epsilon\}$, and
selecting $\epsilon$ is an acknowledged difficulty even there \cite{paes}.
Unsupervised trajectory inference lacks both prerequisites: a labeled loss,
and a scale for $\epsilon$. Three substitutions supply them. The loss is
replaced by the cross-fitted held-out discrepancy of (1), measuring fit
against gene sets excluded during estimation rather than against absent
labels. The fixed threshold becomes a hypothesis test, deriving the set's
width from the empirical dispersion of held-out scores at a stated
significance level. And because testing exact equality would reject any
configuration not identical to $\theta^*$, the evaluation is a
non-inferiority test against a margin $\delta$ estimated from random-seed
replicates, so establishing an empirical scale for
meaningless variation in the data's own units. The resulting acceptance
boundary is
\begin{equation}
\R(\alpha) = \left\{\theta \in \Theta : \theta \text{ well-formed, and }
 T(\Delta(\theta)) \le q_{1-\alpha}\right\},
\end{equation}
where $\Delta_g(\theta) = d_g(\theta) - d_g(\theta^*)$, $T$ is the
block-permuted test statistic, and $q_{1-\alpha}$ is the critical value at
significance level $\alpha$. The set widens where the data constrain the
model weakly and tightens where they constrain it strongly, and its single
tuning parameter is a significance level rather than a magnitude.

Equation (3) is the core modification of this work. Once an admissible set
is constructed, downstream analysis follows standard practice. The claim is
deliberately narrow: that a Rashomon set can be constructed without
ground-truth labels, not that the multiplicity statistic is novel. To the
author's knowledge no prior work defines such a set where the target
quantity is never observed; extensions have moved from classification
through probabilistic classification to regression while relying throughout
on a labeled loss. Hypothesis testing has reached Rashomon sets before, but
for a different question: Paes \emph{et al.}\ \cite{paes} ask whether two
models already inside a loss-defined set can be distinguished on finite
data. That concerns resolution within a set; (3) concerns how the set is
drawn when no loss exists to draw it with.

\subsection{Well-formedness Criteria}
\label{sec:wellformed}

Prior to non-inferiority evaluation, candidate models must satisfy an
intrinsic well-formedness condition.

\begin{itemize}
\item A configuration is deemed well-formed if fewer than 50\% of evaluated
 cells exhibit an exactly uniform posterior fate distribution across all
 terminal states.
\item \textbf{Rationale.} Well-formedness reflects structural validity
 independent of baseline comparisons, analogous to requiring probability
 vectors to sum to one. Because (1) scores cell ordering rather than fate
 assignment directly, a configuration can generate an accurate cell ordering
 while collapsing into a completely uninformative uniform fate
 distribution.
\item \textbf{Concrete case.} A 3-nearest-neighbor configuration achieved an acceptable
 discrepancy score of $d = 0.9921$ (compared to baseline
 $d(\theta^*) = 0.9921$), yet left 99.1\% of cells with an uninformative
 uniform posterior. The well-formedness filter excludes such degenerate
 configurations.
\end{itemize}

\subsection{Module Blocking and Non-inferiority Calibration}

\begin{itemize}
\item \textbf{Co-expression blocking.} Held-out genes are grouped into
 modules via average-linkage hierarchical clustering on
 $1 - |\text{Spearman } \rho|$. Clustered into 15--17 modules per fold (48
 total blocks), blocking prevents correlated gene expression
 within shared transcriptional programs from artificially inflating
 effective sample size and spuriously narrowing $\R(\alpha)$.
\item \textbf{Hypothesis testing.} Non-inferiority tests the null
 hypothesis $H_0 : E[\Delta] \le \delta$, and rejection removes a
 configuration from $\R(\alpha)$. Testing $H_0 : E[\Delta] \le 0$ would
 reject any model non-identical to $\theta^*$ given large sample sizes,
 collapsing $\R(\alpha)$ to a single point.
\item \textbf{Seed-calibrated margin ($\delta$).} The margin $\delta$ is
 estimated empirically by measuring the variation in $D$ across four
 stochastic random-seed refits of the baseline model $\theta^*$, yielding
 $\delta = 1.086 \times 10^{-3}$.
\item \textbf{Sign-flip permutations.} Significance is computed using
 10{,}000 sign-flip permutations across the 48 module blocks (block-resolution limit $2^{-48}$; realized floor $\approx 10^{-4}$ at
10{,}000 draws). Multiple testing across
 hyperparameter configurations is controlled using the
 Benjamini--Hochberg procedure \cite{bh} at $\alpha = 0.05$.
\end{itemize}

\subsection{Label Alignment Across Configurations}

Because unsupervised clustering assigns terminal state labels arbitrarily
across runs, ``Fate 0'' in a candidate configuration may correspond to
``Fate 1'' in the baseline.

\begin{itemize}
\item Candidate output columns are aligned to baseline fates by selecting
 the column permutation that maximizes argmax categorical agreement across
 cells.
\item \textbf{Comparison to mean absolute probability matching.} Aligning
 by minimizing mean absolute probability differences proved unreliable
 because configurations differ in probability saturation
 rates; highly
 saturated runs exhibit numerical offsets under correct alignments that
 caused false label reversals in nine configurations.
\item Argmax matching achieved clear separation (96--100\% agreement for
 true alignments versus 0--4\% for inverted alignments). By selecting the
 labeling layout most similar to baseline, argmax alignment conservatively
 underestimates total predictive multiplicity.
\end{itemize}

\subsection{Synthetic Benchmarking Framework}

Validation against known ground truth utilized synthetic single-cell
trajectories. Expected expression combines a latent progression variable, a
branch assignment, a batch effect, and gene-program membership across seven
programs (shared dynamic, branch-specific, terminal-specific,
transition-specific, housekeeping, batch-affected, and noise), scaled by a
log-normal library size. Counts are drawn from a negative binomial with
dispersion $0.20$, with a Poisson limit when dispersion approaches zero,
followed by count thinning.

\begin{itemize}
\item \textbf{Cohort design.} 75 simulation objects (300 cells, 500 genes
 each) spanning five topology scenarios (clean bifurcation, overlapping
 bifurcation, imbalanced rare branch, continuous non-branching, and
 confounded pseudobranch) across three difficulty levels with five stochastic
 replicates. Difficulty varies five prespecified signal-to-noise
 properties: branch separation, dispersion, count thinning, branch
 imbalance and batch effect strength. It does not vary the topology a
 scenario represents, so an easy and a hard clean bifurcation differ in
 separability but not in what the correct answer is.
\item \textbf{Deterministic generation.} Generated using master seed
 20260808 to allow exact programmatic reconstruction. Ground-truth branch
 labels were isolated from pipeline execution, feature selection, and
 model admission steps.
\item \textbf{Evaluation scope.} Classification error scoring was
 restricted strictly to post-branch cells, as pre-branch cells have
 unresolved developmental fates and inclusion would bound error near 50\%
 regardless of model quality. All statistical tests treat the simulation
 object (replicate), rather than individual cells, as the independent
 experimental unit.
\end{itemize}

\subsection{Validation Against Observed Clonal Fate}

The synthetic cohort validates $\FM$ against a generator, where the correct
fate is a quantity the generator assigned. The lineage-barcoded dataset
permits a stronger test, in which fate was measured in the descendants of
the cell being scored and no part of the label derives from clustering the
expression matrix the prediction is made from. Each clone's fate
composition is the distribution of annotated mature types among its members
at the later timepoints; a state-timepoint cell belonging to exactly one
usable clone inherits it, and cells in more than one clone are dropped as
ambiguous. Of 2{,}081 evaluable cells, 1{,}505 belong to a clone with at
least two fate-timepoint members, which is the subset on which clone purity
carries information, since purity is $1.0$ by construction for a clone with
a single fate member.

Three design points bound the conclusion. The model names its own terminal
states: both adapters support method-inferred terminals only and reject
requested terminal counts outside $\{1,2\}$, reporting that the
implementation ``supports one or two fates'', so the admitted set returns
two unnamed attractors matched to annotated types post hoc by the
orientation achieving greatest agreement. Inference is clustered on clones
rather than cells, since cells of one clone share a label by construction
and a cell-level test overstates its own power; all $p$-values are
2{,}000-replicate bootstraps resampling clones. And the attainable baseline
is not perfect accuracy but the sister ceiling: holding out one fate member
of a clone and predicting it from the remainder gives the accuracy
available with perfect lineage knowledge and no state information.

\subsection{Formal Definitions and Estimation Properties}

For each cell $i$, let $p_{i,k}(\theta)$ represent the predicted
probability of committing to terminal fate $k$ under model $\theta$. The
decision margin under model $\theta$ is
\begin{equation}
m_i(\theta) = p_{i,k^*}(\theta) - \max_{k \neq k^*} p_{i,k}(\theta).
\end{equation}
The certified fate margin is the infimum of the decision margin across all
admitted models in $\R(\alpha)$:
\begin{equation}
\FM_i(\alpha) = \inf_{\theta \in \R(\alpha)} m_i(\theta).
\end{equation}
A cell is certified if $\FM_i > 0$, meaning every admitted model agrees on
the categorical call, and uncertified if $\FM_i \le 0$. The companion
supremum $\bar{m}_i(\alpha) = \sup_{\theta \in \R(\alpha)} m_i(\theta)$
measures upper-bound certainty. On the empirical objects it is
near-saturated, sitting above 1.0 minus a thousandth for more than 97\% of
cells because at least one admitted configuration over-saturates fate
predictions, so it does not separate biological ambiguity from algorithmic
instability there. On the simulation cohort, where saturation is milder, it
instead retains the baseline configuration's discrimination
(Section~\ref{sec:asymmetry}).

Three estimation properties follow. First, computing $\FM_i$ over a finite
grid gives an upper bound on the continuum infimum, so certification is
anti-conservative---a cell certified on a grid may be overturned on a
denser one---while non-certification is strictly conservative. Second,
hypotheses are evaluated once per configuration, since omitting multiplicity
adjustment would let false rejection rates scale with grid size, shrinking
$\R(\alpha)$ and biasing evaluation toward the baseline. Third, the number
of independent blocks $B$ sets the resolution floor for permutation
$p$-values at $2^{-B}$, so an under-blocked test cannot reject anything.
Table~\ref{tab:notation} summarizes the notation.

\subsection{Quantile Certification}
\label{sec:qcert}

$\FM$ as defined above is the $q = 0$ member of a family. For a quantile
level $q \in [0, 1)$, define
\begin{equation}
\FM_i(q) \;=\; \mathrm{Quantile}_q\big\{\, m_i(\theta) \;:\; \theta \in
\R(\alpha) \,\big\},
\end{equation}
and call cell $i$ $q$-certified when $\FM_i(q) > 0$. The claim weakens in a
way that stays interpretable: at $q = 0.25$ over an admitted set of eight,
$q$-certification states that the assignment survives under at least six of
the eight admitted configurations rather than under the single worst one.
Setting $q = 0$ recovers the worst-case definition above exactly. This family is introduced not
as a proposed statistic but as the instrument that localizes where
worst-case certification fails (Section~\ref{sec:aggregator}).

\subsection{Certification Power}

A certified margin is only as meaningful as the model space whose infimum
it is. If every admitted configuration returns nearly the same fate
assignment because the space is narrow rather than because the data are
decisive, $\FM > 0$ is true and uninformative, and the construction cannot
detect this from within: a space of one configuration certifies every cell.
Alongside every multiplicity rate a certification power is therefore reported,

\begin{equation}
\Pi \;=\; \frac{\bar{d}_{\text{space}}}{\bar{d}_{\text{seed}}},
\end{equation}

where $\bar{d}_{\text{space}}$ is the mean pairwise fate-assignment
disagreement between admitted configurations and $\bar{d}_{\text{seed}}$
the mean pairwise disagreement between refits of one configuration
differing only in random seed. The denominator is the irreducible floor.
$\Pi \approx 1$ states that the space disagrees with itself no more than
one configuration disagrees with itself across seeds, so certification over
it asserts nothing beyond seed stability; $\Pi \gg 1$ states that a
surviving margin has survived something.

Both terms require care. The denominator must be computed \emph{within}
each algorithm, since a pair of refits from different algorithms differs by
algorithm rather than seed; 16 of 28 candidate seed pairs on the
haematopoietic object and 8 of 15 on the microglial object are
cross-algorithm and are excluded by construction. Pooling them raises the
floor, depresses $\Pi$, and reverses the direction of
Section~\ref{sec:dilution}. The numerator must be computed over the
admitted set recorded in the ledger, since a differently-sized checkpoint
directory changes it alone.

\subsection{Reproducibility and Analytic Provenance}
\label{sec:provenance}

All analytic choices were fixed in a versioned contract before execution
and verified by hash, so a parameter cannot be altered mid-analysis without
breaking the record. Eight amendments were logged with the observation prompting each. Configurations
that failed to fit are retained in all denominators. The notebook emits a
traceability map linking every numerical assertion to the code stage and
contract hash that produced it, and records an environment lock.

That record is also how the present version differs from an earlier one.
In the configuration grid used for the synthetic cohort, the four
non-seed perturbations were passed to the model constructor twice, once
through the preprocessing configuration and once as a direct keyword, so
every one of them raised at fit time and was silently caught. The admitted
set on all 75 objects was therefore the seed-refit set exactly, $|\R| = 4$
rather than 8, and every simulation-derived quantity in that version
measured reseeding alone. The defect did not touch the three real-data
arms, the clonal validation, $\Pi$, or the multiplicity magnitudes, which
use separate grids. Rerunning the cohort with the grid corrected gives
$|\R| = 8$ on 70 of 75 objects and changes the simulation results
substantially, including the loss of a prespecified positive control
(Section~\ref{sec:aggregator}) and the reversal of the baseline comparison
in Table~\ref{tab:baselines}. Where a corrected value replaces a previously
reported one, both are given. Every quantity derived from the admitted set in
this version is computed by the released library's own routines rather than
reimplemented at the call site, after a reimplementation of the decision
margin in an analysis script was found to disagree with it.

\begin{table}[!t]
\caption{Prespecified Control Configurations and Gate Outcomes}
\label{tab:controls}
\centering\footnotesize
\begin{tabular}{@{}p{0.26\columnwidth}p{0.14\columnwidth}p{0.44\columnwidth}@{}}
\toprule
Configuration & Predicted & Exclusion mechanism \\
\midrule
Scrambled pseudotime & Excluded &
 Failed acceptance test ($p \le 10^{-4}$, stat $+0.0148$) \\
3 nearest neighbors & Excluded &
 Well-formedness (99.1\% uniform) \\
2 principal components & Excluded &
 Well-formedness (54.4\% uniform) \\
50 variable genes & Excluded &
 Pipeline convergence failure \\
Baseline, seed modified & Admitted &
 Passed acceptance test ($p = 0.236$) \\
\bottomrule
\end{tabular}
\end{table}

\section{Results}

\begin{figure*}[t]
\centering
\includegraphics[width=0.68\textwidth]{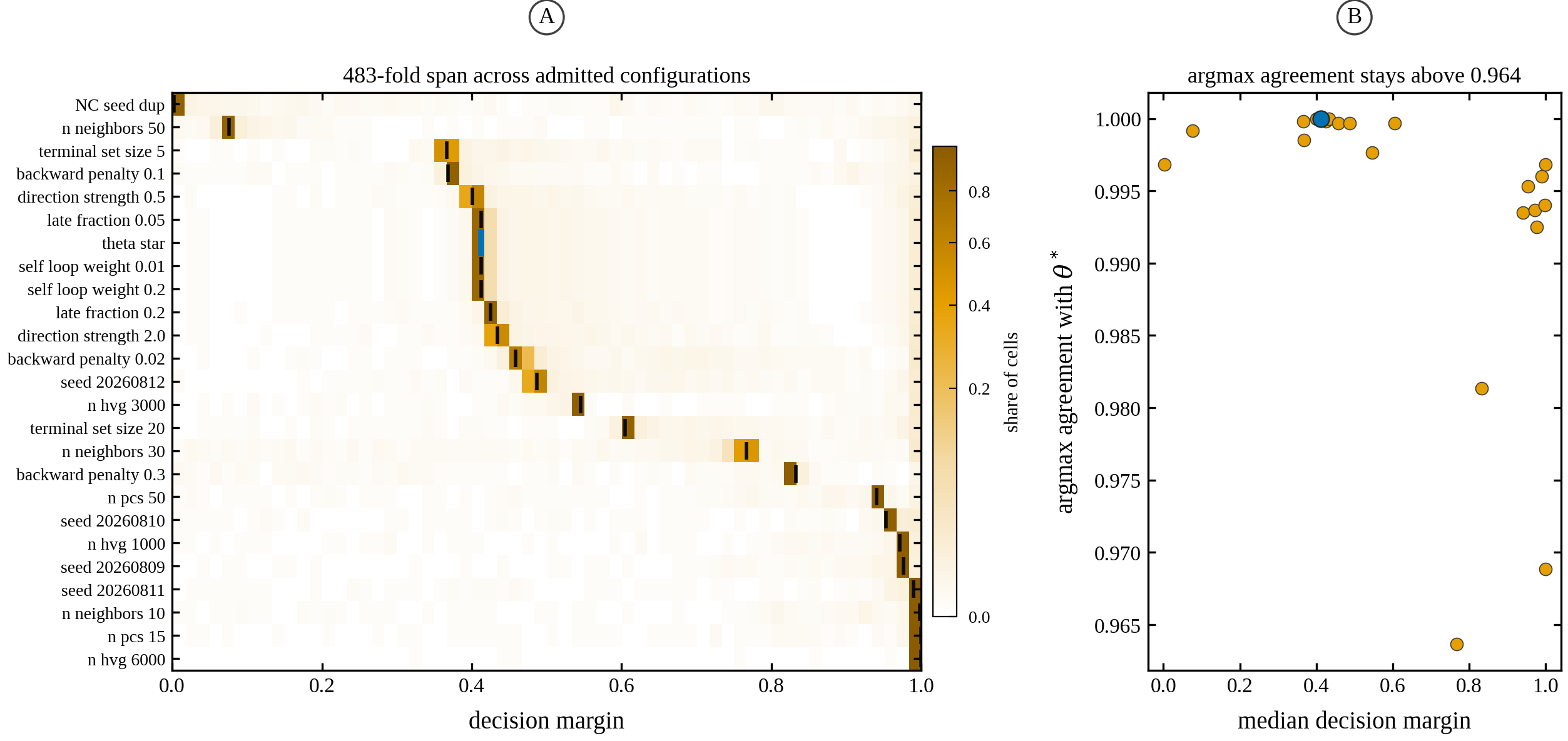}
\caption{Reported confidence varies 483-fold across reconstructions the
data cannot distinguish. (A)~Distribution of the per-cell decision margin under each admitted
configuration, sorted by median. Each row is one configuration, shade gives
the share of its 6{,}000 cells in each margin bin, and the tick marks the
median, with the baseline in blue. (B)~Median margin against argmax agreement with the baseline.
Configurations differ far more in the confidence they report than in the
assignment itself.}
\label{fig:conf}
\end{figure*}

\subsection{Point-estimate Fate Probabilities Display Extreme Parameter Sensitivity}
\label{sec:confspread}

Across statistically indistinguishable parameter settings, reported fate
confidence for individual cells varies up to 483-fold
(Fig.~\ref{fig:conf}). Within any one configuration the distribution of
per-cell margins is narrow (Fig.~\ref{fig:conf}a): almost every cell
receives nearly the same value, and what differs is which value, so
reported confidence is closer to a property of the configuration than of
the cell. A 50-neighbor graph leaves one admitted configuration calling
2.5\% of cells confident (margin above 0.5, defined in (4)); 6{,}000
variable genes leaves another calling 100\% confident. Both are defensible
and both are admitted to $\R(\alpha)$.

The variation is a function of manifold geometry, not of trajectory
inference as such. On the myeloid-progenitor benchmark the identical analysis
spans 0.596 to 0.948, under two-fold, with no admitted configuration
putting more than 2\% of cells above 0.99. The difference is saturation: on
the microglial object three admitted configurations push over 97\% of cells
into extreme certainty while another leaves almost none, so configurations
fall into one regime or the other, and the distribution across them is
correspondingly bimodal near 3--7\% or 99\% confident. On a well-separated
bifurcation this does not occur. How cleanly a manifold isolates its
terminal states is therefore a criterion for when single-model confidence
should be distrusted.

Eleven admitted configurations yield $\Delta_g(\theta) = 0$ exactly, since
they alter only absorbing-walk parameters and leave the scored ordering
unchanged, yet individual fate probabilities shift by up to 0.71 between
these provably equivalent models. Categorical agreement with the baseline
nonetheless stays high, 0.9637 to 1.0000, so at worst an admitted model
reassigns 218 of 6{,}000 cells: sensitivity concentrates in reported
confidence rather than in modal labels.

\subsection{Cross-fitted Discrepancy Evaluates Trajectory Smoothness}

The cross-fitted held-out gene discrepancy metric $D$ differentiates fitted
trajectory reconstructions from perturbed control models. Across held-out genes, 65.5\% exhibit worse
expression smoothness under a scrambled cell ordering, with 23.8\%
degrading by more than two standard deviations relative to their null
distribution.

The baseline scores $D = 0.9921$. Permuting the fitted pseudotime 200 times
and rescoring yields a null distribution centered at $D = 1.0006$ with
standard deviation $0.00053$, placing the baseline 16.0 standard deviations
below the null ($p = 0.005$, the floor attainable with 200 permutations
rather than an estimate of the tail probability). Seed refits track baseline
performance closely ($D = 0.9920$, matching to within $1 \times 10^{-4}$),
while structurally misspecified graphs yield $D = 0.9931$.
Because (2) depends only on rank, no configuration can improve its score by
producing a more widely spread pseudotime axis.

\subsection{Statistical Admission and Evaluation of Prespecified Controls}

All 24 defensible configurations in the one-factor space were admitted;
exclusions arose only among the five prespecified controls registered
before execution (Table~\ref{tab:controls}). The
scrambled ordering was rejected with a statistic of $+0.0148$, an order of
magnitude above the margin. Only that control was excluded by the
acceptance test itself: two were caught by well-formedness and one failed
to fit, and across the 36 configurations spanning both algorithms no
defensible configuration was rejected.

The myeloid-progenitor benchmark shows different mechanics. There the gate
rejected an analytic choice no control was engineered to simulate:
restricting the feature set to 800 variable genes gave a blocked statistic
of $+0.0201$, nearly six times that dataset's margin, at the $p \le
10^{-4}$ floor of 10{,}000 sign-flip permutations. Reducing feature
dimension is a routine choice, so the boundary filters realistic analytic
variation and not only corrupted models---provided the dataset carries
enough variance for the test to resolve differences.

\subsection{Multiplicity Depends More on Model Diversity than on Model Count}
\label{sec:multiplicity}

\begin{figure*}[t]
\centering
\includegraphics[width=0.66\textwidth]{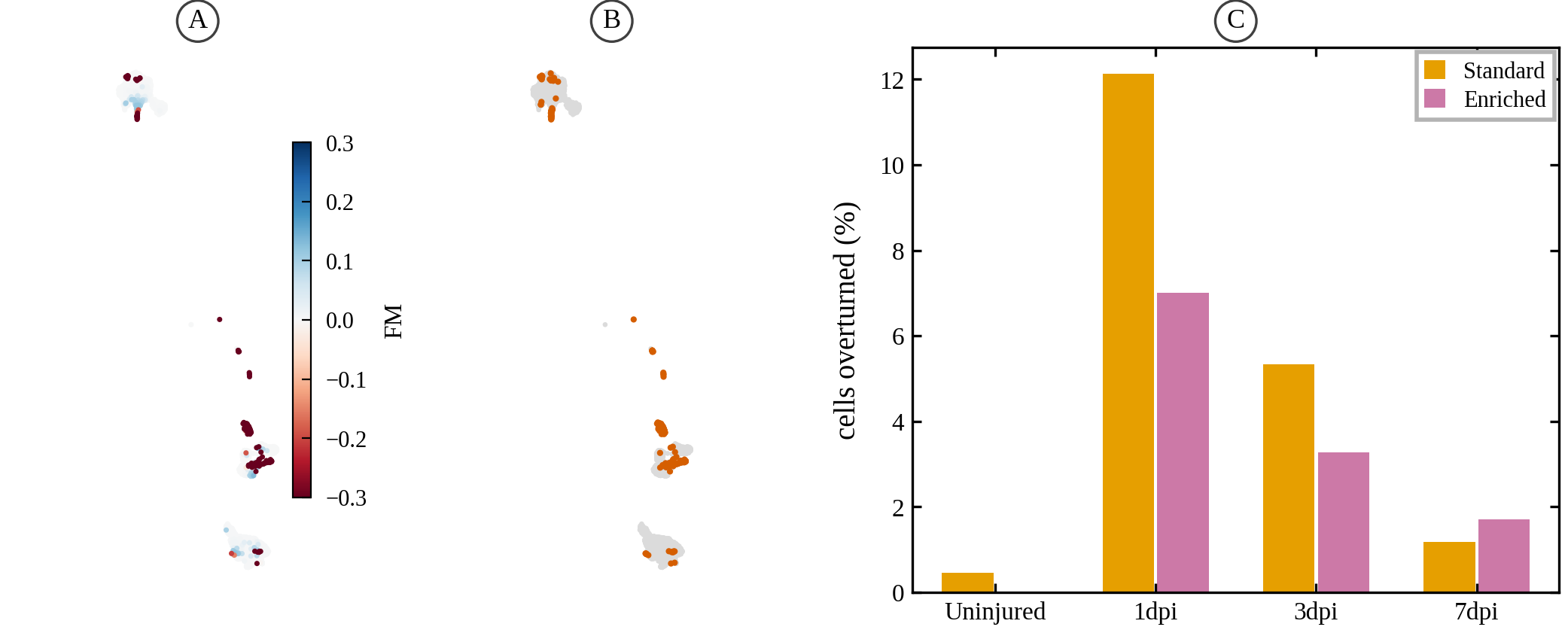}
\caption{Multiplicity is concentrated in the acute injury response.
(A)~$\FM$ across the microglial manifold, embedded by uniform manifold
approximation and projection \cite{umap} from the baseline representation.
(B)~The 235 cells whose assignment is overturned within $\R(\alpha)$; they
cluster rather than scatter. (C)~Fraction overturned by injury timepoint,
split by dissociation protocol; uninjured cord was profiled under one
protocol only.}
\label{fig:where}
\end{figure*}

Evaluating 24 single-factor perturbations within the absorbing-walk grid
yields a baseline multiplicity rate of 3.8\%. The seed replicate registered
as a specificity control (Table~\ref{tab:controls}) also passed and was
admitted as a 25th member, adjusting the rate to 3.9\%, or 235 of 6{,}000
cells; that ensemble is the one the per-cell displays are computed over
(Fig.~\ref{fig:where}). Expanding the candidate space to
include twelve Palantir configurations grows the admissible set to 36 and
raises the rate to 20.4\%.

\begin{figure*}[t]
\centering
\includegraphics[width=0.66\textwidth]{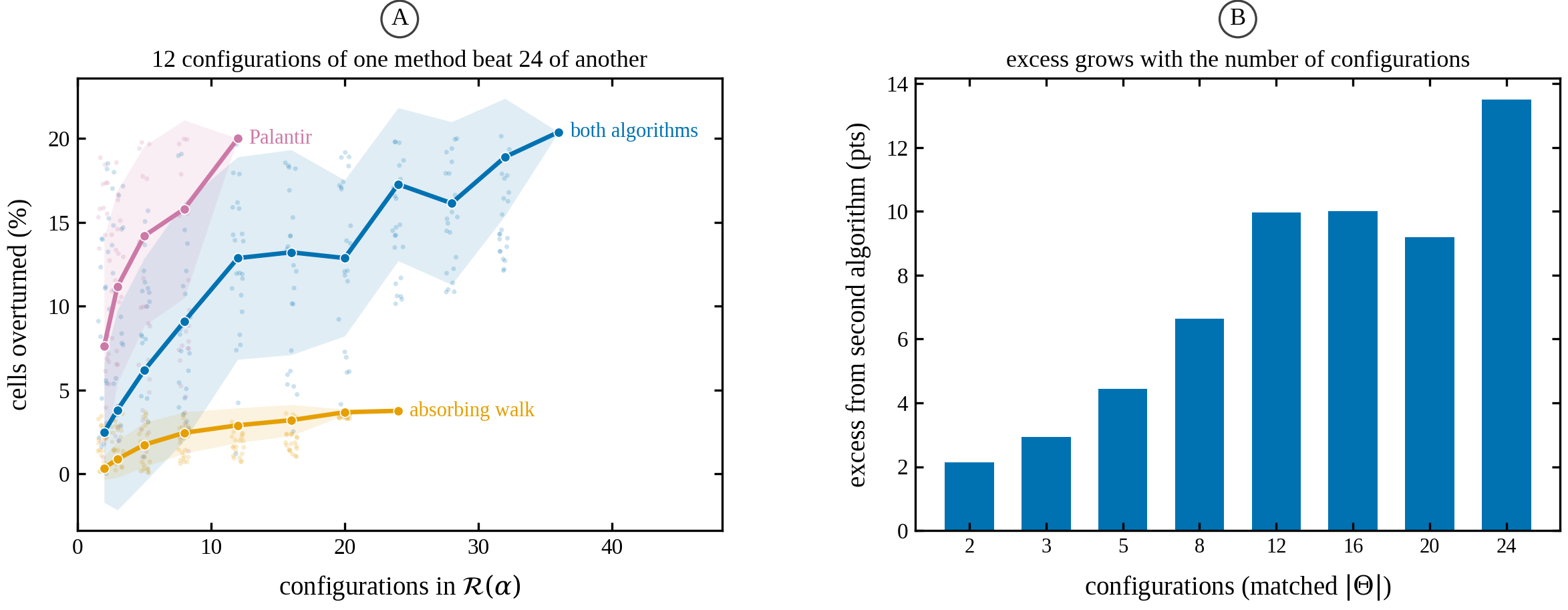}
\caption{Multiplicity depends more on the diversity of the model space than
on its size. (A)~Fraction of cells overturned against the number of
admitted configurations, resampled at each size with bands at $\pm 1$ SD;
the three curves are the pooled two-algorithm set and each algorithm alone. (B)~Excess multiplicity attributable to the second
algorithm at matched candidate-space size, rising monotonically with set
size. Each bar compares two sets of equal size, so the increase is not an
artifact of counting more models.}
\label{fig:space}
\end{figure*}

Resampling the model space at matched cardinality confirms that the
increase is driven by structural diversity rather than model count
(Fig.~\ref{fig:space}a). Sampling 24 configurations restricted to the
absorbing walk leaves the rate at 3.8\%; sampling 24 that span both
algorithms raises it to 17.3\%. The gap widens monotonically with ensemble
size, from $+0.021$ at two configurations to $+0.135$ at twenty-four
(Fig.~\ref{fig:space}b). Most
strikingly, an ensemble of twelve Palantir configurations exposes 20.0\% of
cells where twenty-four absorbing-walk runs expose 3.8\%.

Sensitivity analyses restricted to parameter sweeps within a single
algorithm therefore understate trajectory ambiguity severely, and the
number of configurations evaluated is a misleading summary of how
thoroughly an analytical space has been explored.

Dispersion peaks at an ensemble size of eight (SD 0.071, range 0.007 to
0.203), so two analysts assessing the same number of runs can report
multiplicity figures differing thirtyfold depending on which they sample.
The two-factor expansion of the grid raises the single-algorithm rate to
52.6\%, and 51 of its 96 fitting configurations are rejected as degenerate, almost all combining a backward penalty of 0.3 or
a 50-component projection with a secondary shift---parameter pairs harmless
individually that jointly collapse fate inference. 

Varying the significance threshold $\alpha$ from 0.001 to 0.5 produced no
change in the composition of $\R(\alpha)$ or in the resulting multiplicity
rates in either space. Within the one-factor absorbing-walk grid
the worst admitted statistic was $+0.0015$, against a margin of
$\delta = 1.086 \times 10^{-3}$; the behavior of the boundary across both
algorithms, where one configuration exceeds the margin and is still
admitted, is examined in the repository's admission ledger. Jackknife sensitivity analysis
showed that removing any single configuration altered the one-factor
multiplicity rate by at most 0.005, confirming that the estimate
is not driven by an isolated outlier.

\subsection{Multiplicity Concentrates in Acute Spinal Cord Injury Response}

Overturned cells ($\FM \le 0$) are distributed non-uniformly across the
manifold, clustering at the boundaries of major functional populations
(Fig.~\ref{fig:where}a,b), and their prevalence tracks injury progression:
0.4\% uninjured, peaking at 10.4\% at 1\,dpi, then 4.4\% at 3\,dpi and
1.4\% at 7\,dpi (Fig.~\ref{fig:where}c). Fate assignment is least stable
during the acute phase and settles as the response resolves.

The trend is not a dissociation artifact. It persists within protocol
(12.1\%, 5.3\%, 1.2\% across 1, 3 and 7\,dpi under the standard protocol;
7.0\%, 3.3\%, 1.7\% under the enriched), and sample sizes run counter to a
power artifact, since uninjured tissue has the fewest cells (438) and the
lowest multiplicity while 7\,dpi has the most (2{,}636) and stays low. The
uninjured condition was profiled under one protocol only, so the contrast
both protocols support is 1\,dpi against 7\,dpi.

\subsection{The Multiplicity Rate Reproduces on an Independent Dataset}

To determine whether the preceding observations reflect general properties
of trajectory inference or topological features peculiar to the microglial
manifold, the pipeline was applied without modification to an independent
object of 2{,}730 myeloid progenitors.

On this object the reference configuration $\theta^*$ achieves strong
separation, scoring $D = 0.96685$ against a permutation null of
$1.00079 \pm 0.00120$ ($z = -28.2$, against $z = -16.0$ on the primary
dataset). The estimated noise margin rose to $\delta = 0.00348$ from
$0.00109$, reflecting greater run-to-run variance on the smaller sample.

Of 17 candidate configurations, one was eliminated by the well-formedness
check and one rejected by the non-inferiority gate, leaving 15 admissible.
This corresponds to a multiplicity rate of 3.88\%, closely mirroring the
3.8\% measured on the primary dataset within a single-algorithm
one-factor grid. Across
two biological contexts differing in tissue of origin, cell count and gene
coverage, the rates agree to within 0.1 percentage points.

The confidence spread reported in Section~\ref{sec:confspread} did not replicate.
Comparing the two results separates a general property of the method from a
manifold-dependent one: the frequency with which admissible reconstructions
overturn fate assignments is stable across datasets, while the dispersion
in reported confidence is specific to the underlying developmental
landscape.

\subsection{Disconnect Between Reported Confidence and Certified Robustness}

Comparing baseline reported confidence against certified status
demonstrates that numerical certainty does not guarantee algorithmic
stability. Taking the 75th percentile of baseline confidence (0.708) as a
threshold, 1{,}500 of 6{,}000 cells are classified as confident and 5{,}765
as certified ($\FM > 0$).

However, the two metrics diverge: 201 cells are confident but uncertified,
while 4{,}466 cells are certified but lack high baseline confidence (exact McNemar test, $p < 10^{-15}$). A cell whose baseline confidence is 0.60 can hold the same categorical
assignment across every admitted model, while one at 0.95 can have its fate
call overturned by another admissible configuration. Reported probability and certified margin measure distinct
properties of the pipeline.

\subsection{Certification Power Separates Informative From Vacuous
Certification}
\label{sec:pi}

Table~\ref{tab:pi} reports $\Pi$ for four dataset--algorithm arms spanning
three biological systems.

\begin{table}[t]
\caption{Certification power across four dataset--algorithm arms.
$\bar{d}_{\text{space}}$ is mean pairwise fate-assignment disagreement
between admitted configurations; $\bar{d}_{\text{seed}}$ is the
within-algorithm seed floor. ``Mult.'' is the fraction of cells with
$\FM \le 0$. AUC is of $-\FM$ against clonally observed fate,
which only the barcoded object supplies; the microglial and reprogramming
arms carry no lineage labels and contribute $\Pi$ and multiplicity only.
$\FM$'s AUC against \emph{simulation} truth is 0.682
(Section~\ref{sec:aggregator}) and is not comparable to these, being
measured against a different reference standard on a different cohort.}
\label{tab:pi}
\centering
\small
\setlength{\tabcolsep}{3.5pt}
\begin{tabular}{llrrrrrr}
\toprule
Dataset & Algorithm & $|\R|$ & $\bar{d}_{\text{seed}}$ &
$\bar{d}_{\text{space}}$ & $\Pi$ & Mult. & AUC \\
\midrule
Microglia & Abs.\ walk & 24 & 0.003972 & 0.008595 & 2.16 & 0.038 & --- \\
Haematop. & Abs.\ walk & 15 & 0.1198 & 0.1223 & 1.02 & 0.457 & 0.568 \\
Haematop. & Palantir & 16 & 0.2441 & 0.2791 & 1.14 & 0.847 & 0.615 \\
Reprogram. & Abs.\ walk & 16 & 0.2949 & 0.2437 & 0.83 & 0.743 & --- \\
\bottomrule
\end{tabular}
\end{table}

The microglial arm is the only one whose model space disagrees materially
more than its own seed refits, at 2.2 times the floor. On the two
haematopoietic arms the space disagreement and the seed floor coincide to
within a factor of 1.2, and on the reprogramming arm the space disagrees
\emph{less} than reseeding does, giving $\Pi = 0.83$.

\begin{figure}[t]
\centering
\includegraphics[width=\columnwidth]{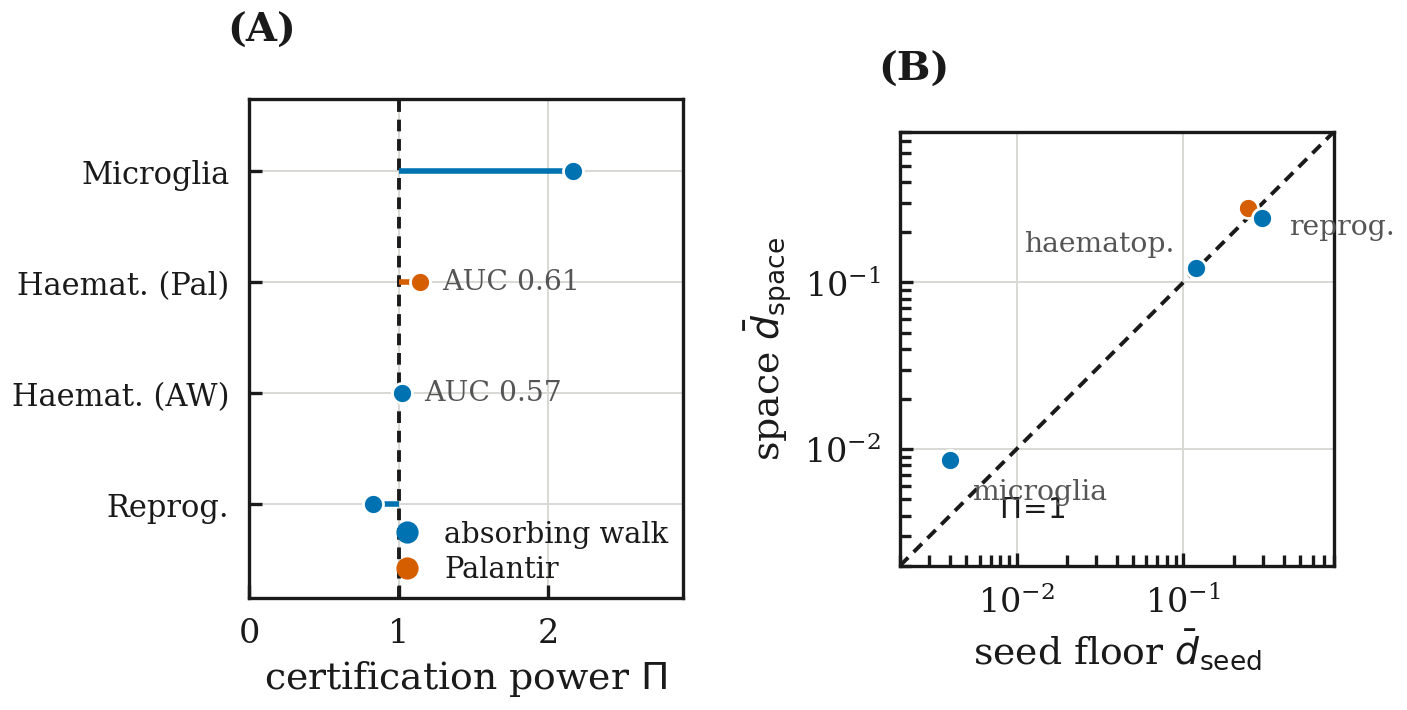}
\caption{Certification power. (A)~$\Pi$ for each of the four
dataset--algorithm arms, with the area under the curve for $-\FM$
predicting clonally observed misassignment annotated on the two arms that
carry lineage labels. Only the microglial arm exceeds $\Pi = 1$
appreciably. (B)~The two components of $\Pi$ per arm,
model-space disagreement against the within-algorithm seed floor, on a
logarithmic axis; an arm on the diagonal has $\Pi = 1$ and certifies
nothing beyond seed stability.}
\label{fig:pi}
\end{figure}

Only the barcoded object supplies fate labels, so $\FM$'s ability to
predict misassignment is measurable on two of the four arms. On those two,
the arm with the higher $\Pi$ also has the higher area under the curve
(0.615 at $\Pi = 1.14$ against 0.568 at $\Pi = 1.02$;
Fig.~\ref{fig:pi}). Two
points agree in order half the time by chance, and this is not presented as an ordering result. The claim $\Pi$ supports is the one it makes by
construction and without labels: a space whose disagreement does not exceed
its own seed floor has certified nothing, and that belongs beside any fate
claim drawn from a certified margin.

This reframes the rates in Section~\ref{sec:multiplicity}: 3.8\% at
$\Pi = 2.16$ and 45.7\% at $\Pi = 1.02$ are not comparable quantities. The
first is measured over a space probing analytic freedom, the second over
one barely exceeding reseeding noise, where a high rate reflects an
unstable algorithm more than a contested biological decision.

\subsection{Adding an Algorithm Can Lower Certification Power}
\label{sec:dilution}

Expanding a model space raises the multiplicity rate monotonically, since
$\FM$ is an infimum, but certification power is not monotone. Holding
cardinality fixed, adding the second algorithm moves $\Pi$ from 2.21 to
0.93 on the microglial object and from 1.03 to 1.46 on the haematopoietic
one, in both cases while the multiplicity rate rises; the discriminating
quantity is the ratio of the two algorithms' own seed floors, 49-fold and
2-fold. Because the denominator of $\Pi$ pools the two algorithms' within-algorithm
seed floors, importing an algorithm whose refits are less stable than the
incumbent's raises that floor faster than it raises space disagreement. A larger
multiplicity rate
obtained this way is therefore not evidence of a better-probed space, and
the rate, $\Pi$ and the per-algorithm floors must be reported together.

\subsection{\texorpdfstring{$\FM$}{FM} Predicts Disagreement With Observed
Clonal Fate}
\label{sec:clonal}

On the fate sub-decision the admitted set's attractors represent, cells the
framework declines to certify disagree with their clone's observed outcome
27.7\% of the time, against 11.3\% among certified cells: a difference of
16.4 percentage points over 1{,}061 scorable cells, with a
2{,}000-replicate bootstrap over clones giving $p < 0.001$. The Palantir
arm on the same cells gives 30.5\% against 19.0\%, a difference of 11.5
points, also at $p < 0.001$. The attainable sister
ceiling for these cells is 0.875.

This addresses the question the construction cannot answer from within.
Agreement on reconstruction of held-out genes is a statement about
pseudotemporal ordering, and whether such agreement carries information
about \emph{fate} is not established by the construction itself; the
discrepancy metric in (1) never sees a fate probability. Here the fate was
measured in the descendants of the cell being scored, by a heritable
barcode, with no part of the label derived from the expression matrix the
prediction was made from. Discrepancy-equivalence does carry fate-relevant
information.

Three qualifications travel with it. The first is that the effect is
measured on the sub-decision the attractors span, and
Section~\ref{sec:topology} reports the ones they do not. The second is that
both haematopoietic arms have $\Pi$ near 1, so this shows $\FM$ tracking
biological error even where certification power is weak, not on a
well-probed space.

The third is that $\FM$ does not outrank the simpler baselines at ranking
individual cells, on either reference standard. Against simulation truth it
is worse than seed dispersion, the single-model margin and the mean margin
(Section~\ref{sec:aggregator}). Against clonally observed fate the same
holds:
on the same cells and labels, seed dispersion reaches 0.662 where $\FM$
reaches 0.568 on the absorbing-walk arm, and the single-model margin
reaches 0.707 where $\FM$ reaches 0.615 on the Palantir arm. The group
difference above is real and significant, but separating two groups by
error rate and ranking individual cells are different tasks, and on
observed fate $\FM$ does the first better than the second. The simulation
comparison does not transfer.

\subsection{Behavior Across Branching Structures}
\label{sec:topology}

The synthetic cohort spans five branching structures: no fork, a false fork
induced by a batch effect, a clean fork, an overlapping fork, and a fork
with one rare branch. Reported per object, with the simulation replicate as
the independent unit, $-\FM$ predicts misassignment with mean area under the
curve 0.61 on the clean fork ($p = 0.275$), 0.72 on the overlapping fork
($p = 0.098$) and 0.73 on the rare-branch fork ($p = 0.027$). Only the last
is nominally significant, and it does not survive correction across the
three. Under an earlier and defective construction of the admitted set these
same three scenarios gave 0.95, 0.91 and 0.94 at $p \le 0.004$
(Section~\ref{sec:provenance}); the apparent topology dependence was an
artifact of that defect. There is no evidence that $\FM$'s informativeness
varies with branching structure, and none surviving multiplicity correction
that its per-cell ordering is informative on any of the three.

Neither the simulator nor either adapter can express a multifurcation: the
first pins the branch count to two for every forking scenario, the second
rejects terminal counts outside $\{1,2\}$. Rather than call a two-branch
cohort a multifurcation, the question is tested where a real one exists. The
barcoded object annotates ten mature types, and the admitted set's two
attractors can be re-scored against every fate pair and one-vs-rest
collapse carrying enough clonally observed cells, giving twenty
sub-decisions across the two algorithms. Of the fifteen the attractors span
at orientation agreement above 0.60, certification is informative in
thirteen, with mean difference 16.4 points and median 11.1, each
significant under the clone bootstrap. One is positive but not significant.
One inverts.

That inversion is worth stating plainly. On the Monocyte-against-rest
collapse under Palantir, at orientation agreement 0.668, uncertified cells
disagree with observed fate 26.6\% of the time and certified cells 45.3\%:
a reversal of 18.7 points at $p < 0.001$, with certified cells reliably
\emph{more} often wrong. Two further inversions fall where orientation
agreement sits at chance, so the attractors do not span the decision and no
information should be expected.

The mechanism is the one already visible on the unbranched simulations. A
certified margin states that the admitted set agrees; if its members share
an attractor structure merging two distinct outcomes, they agree
confidently and identically on the cells that structure mishandles, and
certification tracks the shared structure rather than the truth. A
certified fate claim is therefore conditional on attractor coverage, and
the attractor naming must be reported beside it.

\begin{figure*}[t]
\centering
\includegraphics[width=0.76\textwidth]{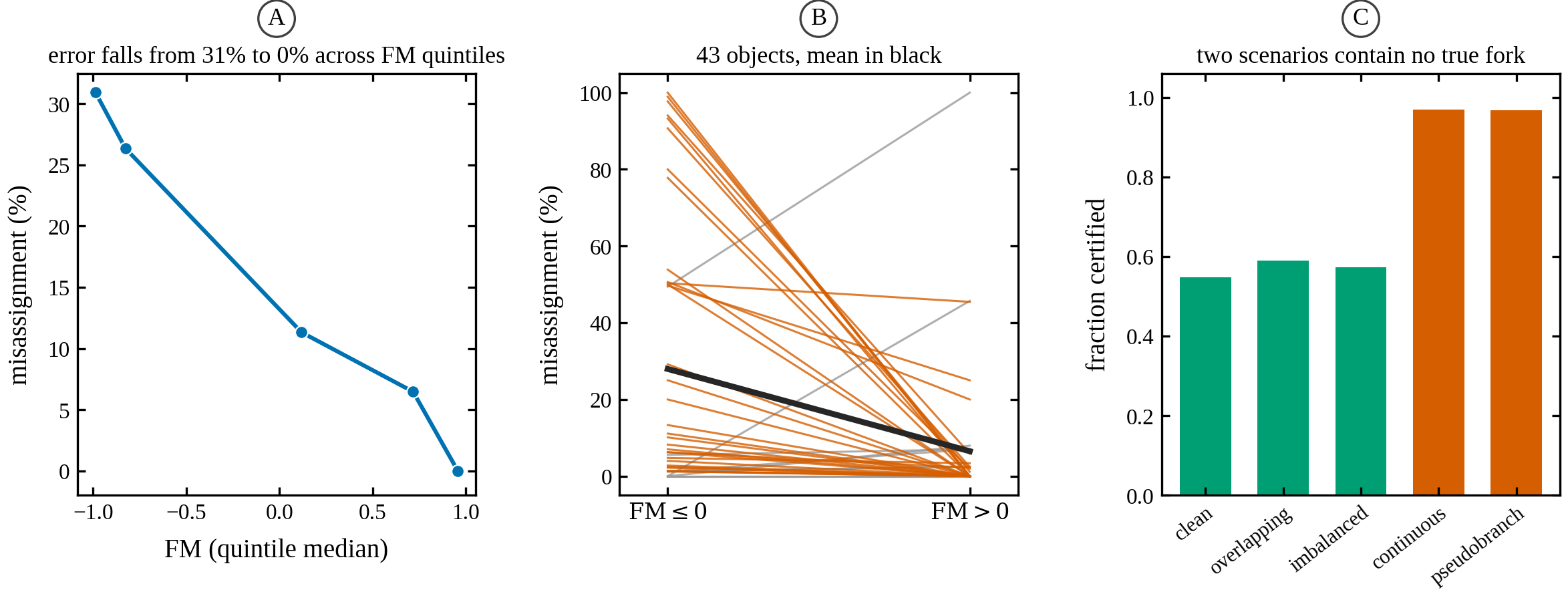}
\caption{Simulation validation on the corrected cohort.
(A)~Misassignment against $\FM$ quintile; error falls from 31\% in the
lowest quintile to 0\% in the highest, so the sign and coarse ordering of
$\FM$ do carry information. (B)~Per-object misassignment either side of
$\FM = 0$, one line per object and the mean in black, over the 43 objects
with cells on both sides. (C)~Fraction certified by scenario. The two
containing no true fork, in red, certify at roughly 0.97---far above the
0.55 to 0.59 of the three that do contain one, which is the inversion
Section~\ref{sec:erosion} addresses.}
\label{fig:sim}
\end{figure*}

\subsection{Certified Fate Margins Do Not Outperform the Fitted Model}
\label{sec:aggregator}

\begin{figure*}[t]
\centering
\includegraphics[width=0.71\textwidth]{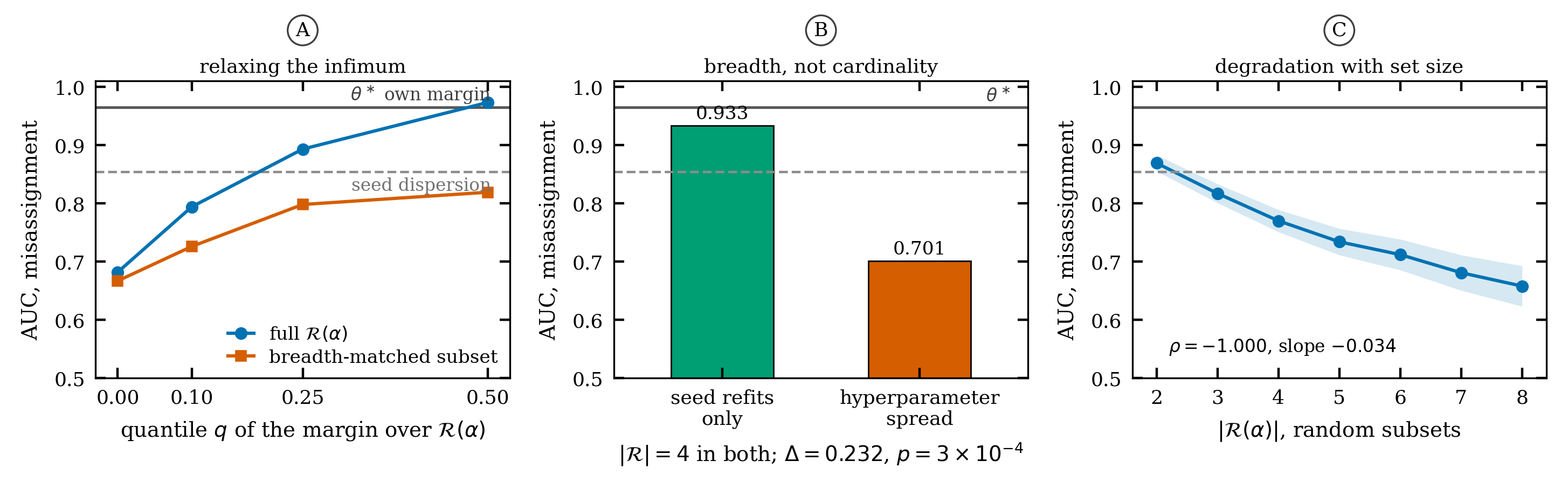}
\caption{Per-cell certification is limited by its aggregator, not by the
quantity being aggregated. Every panel reports mean per-object AUC for
discriminating misassignment against simulation truth over the 28 scorable
forking objects; the solid grey line is the baseline configuration's own
decision margin (0.965) and the dashed line the seed-dispersion baseline
(0.854). (A)~Replacing the infimum over $\R(\alpha)$ with the $q$-quantile
of the same per-model margins. $q = 0$ is $\FM$ as defined. Discrimination
rises monotonically in $q$ but converges toward the single model's own
confidence from below rather than exceeding it. The breadth-matched series
restricts the set to four configurations carrying hyperparameter
perturbations. (B)~At fixed cardinality $|\R| = 4$, four seed refits and
four hyperparameter-perturbed configurations differ by 0.232. Breadth, not
count, is the governing variable. (C)~Discrimination against the size of
randomly drawn admitted subsets, shaded by standard error, falling
monotonically over the seven sizes (Spearman $\rho = -1.000$, slope
$-0.034$ per configuration). The barcoded arm is not drawn: its AUC is
measured against clonally observed fate rather than simulation truth, and
the two are not comparable (Table~\ref{tab:pi}).}
\label{fig:aggregator}
\end{figure*}

$\FM$ was introduced to give the analyst a per-cell reliability statistic
that a single fitted model cannot. Evaluated against simulation ground truth
it does not deliver one, and this section reports that result together with
the diagnostics that localize the failure.

Across the corrected cohort, misassignment is 28.5\% among uncertified cells
($\FM \le 0$) against 4.9\% among certified cells ($\FM > 0$), and falls
monotonically from 31\% in the lowest $\FM$ quintile to 0\% in the highest, so the sign of $\FM$ does separate error. The ordering it induces, however, is weak.
Taking the simulation replicate as the independent unit over the 28 forking
objects that pass every scoring filter, $-\FM$ discriminates misassignment
at mean AUC 0.682. A prespecified positive control does not survive:
$\FM$ now correlates with true geodesic distance from the branch point at
Spearman $\rho = 0.030$, against 0.399 under an earlier and defective
construction of the admitted set (Section~\ref{sec:provenance}). Cells
further downstream of a fork are not meaningfully more certifiable.

\begin{table}[!t]
\caption{Discrimination of Misassignment Against Simulation Truth, Every
Statistic Computed on the Same Cells}
\label{tab:baselines}
\centering\footnotesize
\begin{tabular}{@{}p{0.52\columnwidth}rr@{}}
\toprule
Predictor & AUC & vs.\ $\theta^*$ \\
\midrule
Margin at $\theta^*$ (single model) & 0.965 & --- \\
$q = 0.50$ margin over $\R(\alpha)$ & 0.973 & $p = 0.221$ \\
$q = 0.25$ margin over $\R(\alpha)$ & 0.893 & $p = 0.367$ \\
Mean margin over $\R(\alpha)$ & 0.883 & $p = 0.374$ \\
$\bar{m}$, the supremum over $\R(\alpha)$ & 0.973 & $p = 0.494$ \\
Seed dispersion (no acceptance region) & 0.854 & $p < 0.001$ \\
Bootstrap stability & 0.681 & $p < 0.001$ \\
$\FM$, the infimum over $\R(\alpha)$ & 0.682 & $p = 0.003$ \\
\bottomrule
\end{tabular}

\vspace{3pt}
\footnotesize Mean of per-object AUC over the 28 forking objects passing all
scoring filters. The right column gives the Wilcoxon signed-rank $p$ for the
paired per-object difference against the margin at $\theta^*$ (zeros
dropped, normal approximation; the exact test is invalid under ties).
Nothing beats the margin at $\theta^*$. The supremum $\bar{m}$, the median
margin, the mean margin and the $q = 0.25$ margin are indistinguishable
from it; seed
dispersion, bootstrap stability and $\FM$ are all significantly worse, $\FM$
by 0.284. Seed dispersion computed from the probabilities and from the
margins is one ranking, not two, since the margin is $2\pi_{i,k^*} - 1$ for
a two-fate decision, and only the former is listed. Values are not
comparable with those in an earlier version of this work,
which were computed over a defective admitted set and pooled over cells
rather than averaged over objects.
\end{table}

\begin{figure*}[t]
\centering
\includegraphics[width=0.76\textwidth]{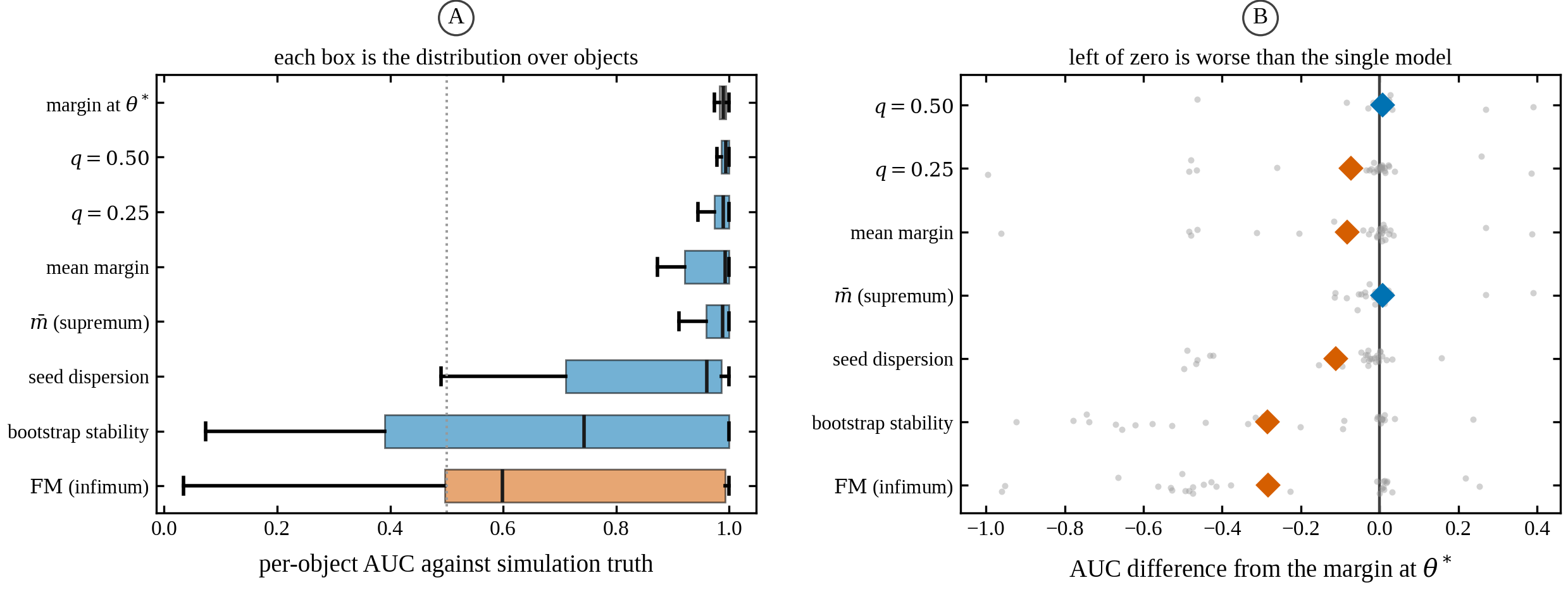}
\caption{The baseline comparison per object, over the 28 forking objects
passing all scoring filters. Table~\ref{tab:baselines} gives the means;
this gives the distributions behind them.
(A)~Per-object AUC for each predictor; the dotted line marks chance. The
margin at $\theta^*$, the $q = 0.50$ margin and the supremum $\bar{m}$ sit
tight against the ceiling, while $\FM$ and bootstrap stability spread down
to near zero on individual objects. (B)~Each predictor's AUC difference
from the margin at $\theta^*$, one grey point per object, mean as a
diamond. Points left of zero are objects where the predictor is worse than
the single fitted model.}
\label{fig:bench}
\end{figure*}

\subsubsection{Comparison against the alternatives}
Trajectory software already reports a fate probability, and an analyst
worried about stability has cheaper options than a new construction. $\FM$
is compared against four on the same cells: the reported margin at
$\theta^*$; the companion supremum $\bar{m}$; the dispersion of the decision
margin across reseedings of $\theta^*$, which needs no acceptance region at
all; and bootstrap stability, refitting $\theta^*$ on 20 subsamples
retaining 80\% of cells and counting label changes. For a two-fate decision
confidence, margin and entropy are monotone transforms of one another and
count as one predictor, and the seed dispersion of the probabilities and of
the margins are likewise the same ranking, since the margin is
$2\pi_{i,k^*} - 1$.

$\FM$ loses to every one of them except bootstrap stability, which it ties
(Table~\ref{tab:baselines}, Fig.~\ref{fig:bench}). The decisive comparison
is against
the margin at $\theta^*$: $\FM$ is worse by 0.284, in 18 of 28 objects,
$p = 0.003$. Nothing constructed over the admitted set beats the single
fitted model.

\subsubsection{The upper tail does not fail, and that is informative}
\label{sec:asymmetry}
The supremum $\bar{m}$ reaches 0.973 and is statistically indistinguishable
from the margin at $\theta^*$ ($+0.008$, $p = 0.49$), while the infimum over
the same set reaches 0.682. Two order statistics of one collection of
margins, 0.29 apart. The asymmetry has a cause, and it narrows the claim
this paper can make.

Because $\theta^*$ is itself a member of $\R(\alpha)$ and $k^*$ is
$\theta^*$'s own assignment, $m_i(\theta^*) \ge 0$ for every cell, and
therefore
\begin{equation}
\bar{m}_i(\alpha) \;\ge\; m_i(\theta^*) \;\ge\; \FM_i(\alpha).
\end{equation}
The supremum is bounded below by the baseline model's own margin and is
close to a monotone transform of it over most cells, so it inherits
$\theta^*$'s signal rather than measuring the model space. The infimum has
no such floor: it is free to run to whichever admitted configuration
disagrees most, and nothing anchors it to the cell. The defect is therefore
specific to the lower tail of the margin distribution over $\R(\alpha)$, not
a general property of order statistics, and a worst-case certificate is
exactly the statistic that reads that tail.

\subsubsection{The failure is in the aggregator}
Three diagnostics locate the defect, and they rule out the explanations that
would make it a property of this dataset.

\emph{Breadth, not cardinality.} Admitted sets of identical size behave very
differently depending on what their members disagree about. At $|\R| = 4$,
four random reseedings of $\theta^*$ give 0.933, while four configurations
carrying hyperparameter perturbations give 0.701---a matched difference of
0.232 in 20 of 28 objects, Wilcoxon $p = 3\times10^{-4}$
(Fig.~\ref{fig:aggregator}b). Discrimination also falls monotonically with
the size of randomly drawn subsets, from 0.869 at $|\R| = 2$ to 0.658 at
$|\R| = 8$ (Spearman $\rho = -1.000$ over the seven sizes, slope $-0.034$
per configuration; Fig.~\ref{fig:aggregator}c). That curve should not be read
as a size law: larger random subsets of a fixed eight-configuration pool
necessarily draw in more hyperparameter perturbations, so size and breadth
are confounded by construction, and the fixed-cardinality contrast shows
breadth carries more of the effect than the entire size range does.

\emph{Not saturation.} An infimum could lose discrimination simply by
driving cells onto a common floor and tying them. It does not. The fraction
of tied cell pairs is 0.010 at $|\R| = 2$ and 0.006 at $|\R| = 8$, falling
rather than rising, and only 4.6\% of cells lie within $10^{-6}$ of their
object's minimum at $|\R| = 8$. On the balanced panel of objects scorable at
every subset size, discrimination restricted to cells still certified is
flat in $|\R|$ (Spearman $\rho = -0.071$, $p = 0.88$). Where the floor has
not been reached, the ordering is intact.

\emph{Relaxing the infimum recovers the signal, up to a ceiling.} If the
aggregator is the defect, the per-model margins should still carry per-cell
information. Replacing $\inf$ with the $q$-quantile of the same margins
gives 0.682, 0.794, 0.893 and 0.973 at $q = 0$, 0.10, 0.25 and 0.50 over the
full admitted set (Fig.~\ref{fig:aggregator}a), monotone in $q$, with
$q = 0.10$ exceeding the infimum by 0.113 in 19 of 28 objects
($p = 3\times10^{-4}$). The information is present; the infimum discards it.
But the recovery converges to the margin at $\theta^*$ rather than past it.
At $q = 0.50$ the difference against the single model is $+0.007$,
$p = 0.22$. Relaxing the certificate is not a repair, it is an interpolation
back toward trusting one model's probability.

Taken together with Section~\ref{sec:asymmetry}, these say that the
infimum over a model set is an \emph{unanchored} extreme order statistic.
Over near-replicate members it tracks genuine per-cell fragility; as the set
broadens it is set by whichever single admitted configuration disagrees
most, and ceases to describe the cell at all. The supremum over the same
margins does not behave this way, because $\theta^*$'s own membership
bounds it from below. This is a property of the construction, not of this
pipeline or this cohort, and it should be expected wherever a worst-case
per-point certificate is taken over a Rashomon set assembled from
qualitatively heterogeneous models.

\subsubsection{What this does not overturn}
The admitted set still carries fate-relevant information; the result above
concerns one way of summarizing it. Section~\ref{sec:clonal} shows that
against clonally observed fate, cells the framework declines to certify
disagree with their clone's outcome 16.4 points more often than certified
cells. Multiplicity magnitude (Section~\ref{sec:multiplicity}),
certification power, margin erosion (Section~\ref{sec:erosion}) and the
sub-decision analysis are all independent of the per-cell ranking that fails
here, and none of them is affected.

\subsection{Margin Erosion Distinguishes Real From Spurious Branch Points}
\label{sec:erosion}

Certification alone cannot separate a real fork from an arbitrary partition
of continuous data: the unbranched scenarios certify at roughly 97\%, far
above the 55--59\% of any genuine fork (Fig.~\ref{fig:sim}c). The ratio of the certified margin to the baseline
mean margin measures how far the model space erodes that margin, and
separates the two cases where certification does not, with an area under the
curve of 0.890 (Mann--Whitney $p = 6.2 \times 10^{-9}$) across the 45 forking
and 30 non-forking objects. The scope is narrow and should be read as such---75
synthetic objects, one algorithm, no-fork controls from the generator that
made the forks, and eight forks and nine non-forks misclassified at the
best threshold. It is evidence that a structural control is
\emph{constructible} from quantities the framework already computes, not a
validated test, and it is untested on real data, since no empirical arm
carries a known no-fork control. 

\section{Discussion and Limitations}

\subsection{Methodological Scope and Interpretation Boundaries}

Validation on synthetic datasets containing no true lineage branch points
reveals a key theoretical limitation of label-free multiplicity frameworks.
In simulations generating continuous non-branching progressions or
batch-confounded pseudobranches, certification reaches 96.9\% and 96.8\%,
against 55\% to 59\% on the three scenarios that contain a true fork.

Because absorbing random walks force continuous inputs into discrete
terminal states, every admitted configuration partitions unbranched data in
a similar, arbitrary way. High agreement across $\R(\alpha)$ then reflects
shared algorithmic inductive bias rather than biological ground truth: it
confirms that an assignment is analytically determined within a model
class, not that a lineage bifurcation exists.

The consequence extends past trajectory inference. When an algorithm class
shares a strong inductive bias, every member of the admissible set inherits
it, and predictive multiplicity becomes insensitive to structural
misspecification. Low multiplicity can arise from a shared blind spot as
readily as from genuine signal, so any Rashomon-set method applied to a
family with a strong common prior faces the same failure, and reporting low
multiplicity without a structural control invites the misreading. The
existence of the fate structure is an assumption of the construction rather
than an output of it, and certification alone does not verify it. That
assumption is not beyond the reach of quantities the framework already
computes: the margin-erosion ratio of Section~\ref{sec:erosion} separates
genuine forks from imposed partitions at AUC 0.890, where raw certification
orders the two cases backwards. It is offered as a direction rather than a
validated control, since its no-fork cases come from the same generator as
its forks and it is untested on real data.

\subsection{Limitations of the Discrepancy Metric}
\label{sec:dlimits}

Three constraints follow from scoring ordering rather than fate. Equation
(1) evaluates cell ordering, so a configuration that orders well while
collapsing fate probabilities passes the discrepancy gate and must be
caught by the well-formedness screen (Section~\ref{sec:wellformed}). Baseline discrepancy
on empirical data ($D = 0.9921$) sits close to the permutation null
($D = 1.0006$), so absolute thresholds cannot transfer across datasets
without recalibration. And the metric needs gene depth: on a low-yield
object of 618 Schwann cells, an ordering fitted directly to the held-out
genes---a circular procedure no legitimate reconstruction can exceed---reached only $D = 0.936$ against a null of 1.000, because 202
well-detected genes against 12{,}287 in microglia compresses the dynamic
range past the point where models separate.

\subsection{Certification Is Conditional on Attractor Coverage}

The unbranched simulations show certification near 97\% where no fork
exists, against 55--59\% where one does. Section~\ref{sec:topology} shows the same mechanism on real data
against clonally observed fate, which is the stronger form of the
observation: on a fate sub-decision the admitted set's attractors do span,
certified cells disagree with observed fate 18.7 percentage points
\emph{more} often than uncertified cells.

Both share one cause. $\FM$ certifies that every admitted configuration
assigns a cell the same fate. Where the set shares an attractor structure
merging two genuinely distinct outcomes, its members agree confidently and
identically on exactly the cells that structure mishandles, and the
surviving margin measures the agreement of a shared error rather than the
determinacy of a correct answer. Three practices follow: a certified fate
claim should carry the attractor naming it rests on, certification should
be reported with $\Pi$, and low multiplicity should not be read as evidence
of biological branching.

\subsection{The Two-State Fate Ceiling}

Both adapters reject requested terminal counts outside $\{1,2\}$, each
reporting that the implementation supports one or two fates while
advertising multi-fate support among its declared capabilities; the
simulator is bound equivalently. Every fate result here is therefore a
two-way decision, including those drawn from an object annotating ten
mature types. This is a boundary of the available implementations rather
than of the construction, appearing as a single guard in each adapter, and
its consequence is that the behavior of $\FM$ under genuine multifurcation
is addressed only through sub-decisions of a real one.

\subsection{Limitations and Assumptions}

This study evaluated four empirical datasets and two algorithms over
discrete hyperparameter grids. The two algorithms share a
\texttt{graph\_diffusion} provenance family, so the multiplicity increment
from expanding beyond one algorithm is a lower bound rather than a measured
cross-family effect; no implementation spanning a structurally different
family was available within the frozen environment. Evaluating a discrete grid captures
practical non-identifiability across user choices rather than formal
mathematical non-identifiability in continuous parameter limits. The
6{,}000-cell subsampling represents a computational memory limit for dense
matrix calculations, and synthetic benchmarks utilized 300-cell objects.
Clonal lineage labels are available for one of
the four datasets (Section~\ref{sec:clonal}) and absent in the other three, so the
fate-validity claim rests on that dataset while the multiplicity and
multiplicity results rest on all four arms and the certification-power
results on the three datasets they cover.

\subsection{Future Work}

The negative result of Section~\ref{sec:aggregator} sharpens rather than
closes the question, and three extensions follow directly from it.

The first is a model space spanning more than one algorithm family. Both
algorithms evaluated here report a \texttt{graph\_diffusion} provenance, so
every cross-algorithm multiplicity increment above is a lower bound. Since
breadth is what degrades the infimum, a space including an optimal-transport
or RNA-velocity estimator would be the sharpest available test of the
mechanism: it predicts that worst-case certification gets worse, not better,
as the space becomes more representative.

The second is whether any aggregator over an admitted set beats the fitted
model. The quantile family of Section~\ref{sec:qcert} converges to the
single model's confidence from below; whether a statistic that weights
admitted configurations by their discrepancy, rather than treating them as
exchangeable, can exceed it remains open.

The third is a fate model admitting more than two terminal states. The
ceiling is an implementation boundary, not a property of the construction:
the margin generalizes to the gap between the top two of $K$ probabilities,
though a cell may hold a wide top-two gap while remaining uncertain among
the remainder.

\section{Conclusion}

By replacing supervised loss tolerances with seed-calibrated
non-inferiority testing on held-out genes, FateMultiplicity quantifies
predictive multiplicity in unsupervised trajectory inference without
ground-truth labels. Measuring it is worthwhile and the magnitudes are
large. Model diversity drives observed ambiguity far more than
configuration count: twelve runs of one algorithm expose more multiplicity
than twenty-four of another, and since both share a provenance family that
increment is a lower bound. The multiplicity rate is consistent across
datasets within a single-algorithm one-factor grid (3.8\% and 3.9\%) while the dispersion
in reported confidence is not, spanning 483-fold on one manifold against
under two-fold on the other. Multiplicity is not uniform across a tissue: in
injured spinal cord it peaks during the acute phase, when transcriptional
state changes fastest. And a rate is uninterpretable without its
certification power---across four arms $\Pi$ spans 2.16 to 0.83, only one
space disagreeing with itself appreciably more than reseeding does, so rates
from different arms are not comparable quantities.

Certifying individual cells against that set is a different matter, and the
central negative result of this paper is that it does not work. Against
simulation ground truth $\FM$ discriminates misassignment worse than the
fitted model's own decision margin, by 0.284 AUC at $p = 0.003$, and worse
than a seed-dispersion baseline that requires no acceptance region. The
cause is the aggregator rather than the quantity. An infimum over a model
set is an extreme order statistic: over near-replicate members it tracks
per-cell fragility, and as the set broadens it reports whichever admitted
model is most extreme instead. Breadth, not cardinality, governs the
degradation; it is not tie saturation; and relaxing the infimum to a
quantile recovers the discarded signal monotonically but converges toward
the single model's own confidence rather than exceeding it. This should be expected
wherever worst-case per-point certification is attempted over a Rashomon set
assembled from heterogeneous models, and it is reported in full because the
construction is an attractive one that others are likely to try.

What the admitted set does support is coarser and still useful. Against
clonally observed fate, cells the framework declines to certify disagree
with the measured outcome of their clone 16.4 percentage points more often
than cells it certifies, so discrepancy-equivalence carries fate-relevant
information even though the per-cell ranking built from it does not beat the
baseline. A margin-erosion ratio separates real forks from imposed
partitions of continuous data at AUC 0.890 in simulation, untested on real
data, where certification alone orders the two cases backwards. And certification remains conditional on attractor
coverage: where the shared algorithmic structure merges two real outcomes,
certified cells can be reliably more often wrong than uncertified ones.

The practical recommendation follows from the negative result rather than
despite it. Report the multiplicity rate and the certification power
alongside point-estimate fate probabilities, name the attractors any fate
claim rests on, and treat low multiplicity as evidence of analytical
determinacy within a model class rather than of biological branching. Do not
replace the fitted model's own fate probability with a worst-case
certificate over an admissible set; on the evidence here it is the weaker
statistic, and the broader and more honestly constructed the set, the weaker
it gets.

\section*{Data and Code Availability}

All four datasets are public: GEO accessions GSE162610, GSE72857 and
GSE99915, and the lineage-barcoded haematopoietic data of \cite{weinreb}.
Simulation objects are regenerated deterministically from the recorded
master seed (20260808) rather than archived as matrices.

Code is available at \url{\repourl}. The analysis reported here is
\texttt{FateMultiplicity.ipynb}, which executes the full pipeline and emits
the versioned contract, the gene-partition and module registry, the
per-configuration admission ledger, and the environment lock.
\texttt{FateMultiplicityPrework.ipynb} documents the dataset selection and
discrepancy-function development described in Section~\ref{sec:dlimits}. The repository
also archives the notebooks of two preceding studies by the same author,
which are not part of this work: \texttt{FateStability.ipynb}, a trajectory
audit from which the simulation cohort and the perturbation axes are
inherited, and \texttt{Paper.ipynb} with
\texttt{Additional\_Validation.ipynb}, an earlier analysis of peripheral
and central nervous system injury that supplied the object used for the
ceiling test in Section~\ref{sec:dlimits}.

\section*{Declarations}

Arjun Bhupatiraju performed conceptualization, methodology, analysis,
visualization and writing. The work received no external funding, and the
author declares no relevant financial or non-financial competing interests.
Institutional review board and animal ethics review were not required, as
this study is a secondary analysis of publicly available mouse sequencing
data and involved no new human participants or animal experiments.

\section*{Acknowledgments}

The author wrote the manuscript. An AI assistant (Claude, Anthropic) was
used for analysis code development, methodological discussion, figure
generation, and editorial review of the manuscript text. The author conceived the study, designed the analytic protocol and its
gates, selected the datasets, reviewed and approved every analytic
decision, determined which results stood and which required revision, and
verified every reported result against the recorded evidence chain. Sole
responsibility for the content rests with the author, Arjun Bhupatiraju,
and not with the assistant.


\end{document}